\documentclass{article}

\usepackage{arxiv}

\usepackage[utf8]{inputenc} 
\usepackage[T1]{fontenc}    
\usepackage{hyperref}       
\usepackage{url}            
\usepackage{booktabs}       
\usepackage{amsfonts}       
\usepackage{nicefrac}       
\usepackage{microtype}      
\usepackage{lipsum}
\usepackage{graphicx}
\usepackage{siunitx}
\usepackage{subcaption}

\newcommand\blfootnote[1]{%
  \begingroup
  \renewcommand\thefootnote{}\footnote{#1}%
  \addtocounter{footnote}{-1}%
  \endgroup
}

\title{Pre-interpolation Loss Behaviour in Neural Networks}

\author{
  Arthur E.W.~Venter, Marthinus W.~Theunissen, Marelie H.~Davel\\
  Multilingual Speech Technologies, North-West University, South Africa; and CAIR, South Africa.\\
  \texttt{aew.venter@gmail.com, tiantheunissen@gmail.com, marelie.davel@nwu.ac.za} \\
  
}

\begin{document}
\maketitle
\blfootnote{The final authenticated publication is available online at \url{https://doi.org/10.1007/978-3-030-66151-9_19}}

\begin{abstract}

When training neural networks as classifiers, it is common to observe an increase in average test loss while still maintaining or improving the overall classification accuracy on the same dataset.  In spite of the ubiquity of this phenomenon, it has not been well studied and is often dismissively attributed to an increase in borderline correct classifications. We present an empirical investigation that shows how this phenomenon is actually a result of the differential manner by which test samples are processed. In essence: test loss does not increase overall, but only for a small minority of samples. Large representational capacities allow losses to decrease for the vast majority of test samples at the cost of extreme increases for others. This effect seems to be mainly caused by increased parameter values relating to the correctly processed sample features. Our findings contribute to the practical understanding of a common behaviour of deep neural networks. We also discuss the implications of this work for network optimisation and generalisation.

\end{abstract}


%
%
\section{Introduction}
\label{sec:intro}

According to the principal of \textit{empirical risk minimisation}, it is possible to optimise the performance on machine learning tasks (e.g. classification or regression) by reducing the empirical risk on a surrogate loss function as measured on a training dataset \cite{GBC16}. The success of this depends on several assumptions regarding the sampling methods used to obtain the training data and the consistency of the risk estimators \cite{murphy}. Assuming such criteria are met, we expect the training loss to decrease throughout training and that the loss on samples not belonging to the training samples (henceforth referred to as validation or evaluation loss) will initially decrease but eventually increase as a result of overfitting on spurious features in the training set.

\begin{figure}[!h]
        \centering
        \includegraphics[width=0.98\linewidth]{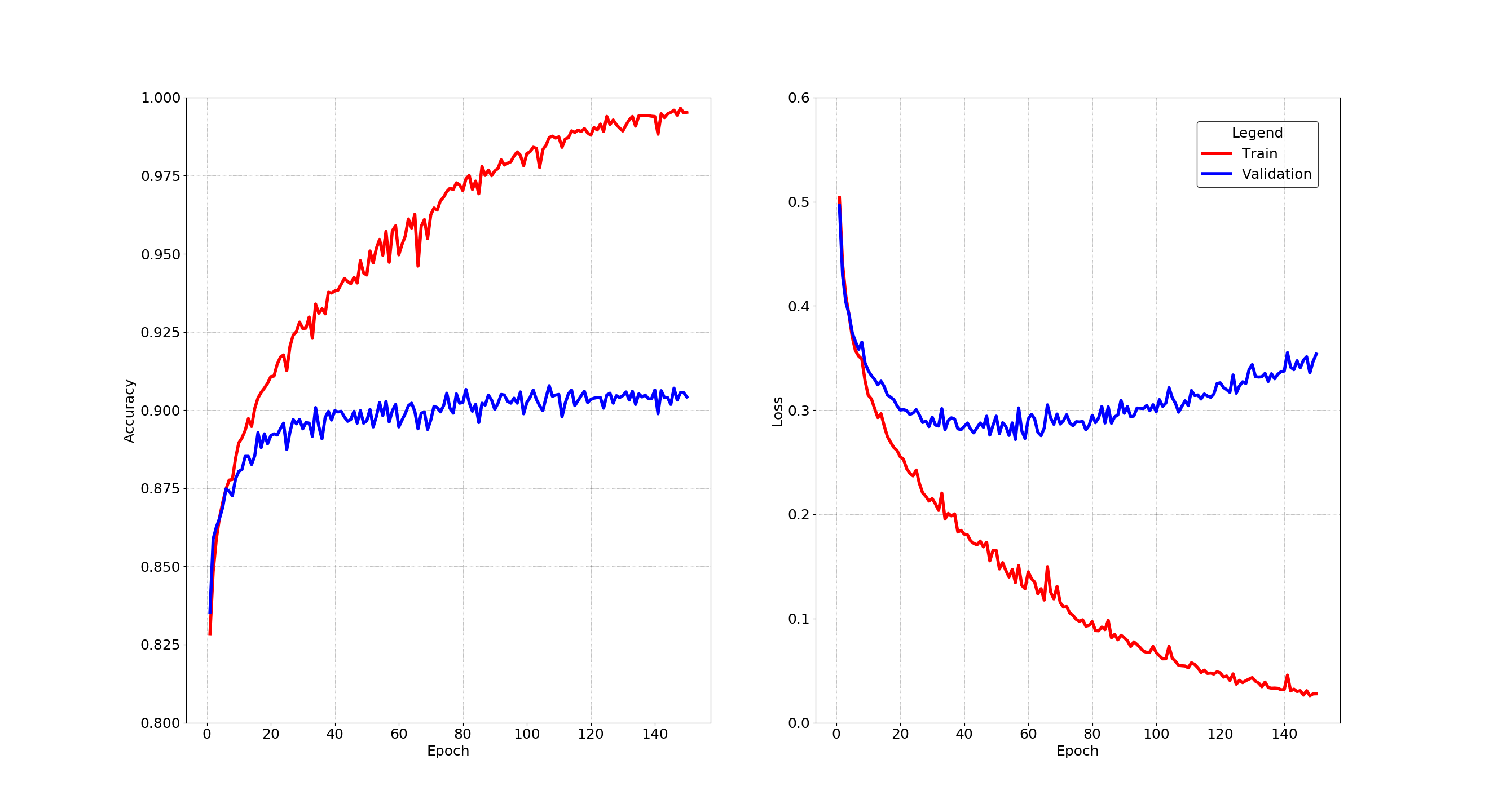}
        \caption{Learning curves of an example $1 \times 1\ 000$ MLP trained on $5\ 000$ FMNIST samples using SGD with a mini-batch size of $64$. The network clearly shows an increasing validation loss with a slightly increasing validation accuracy.}
        \label{fig:Introductory_Example}
\end{figure}

Actual performance is usually not directly measured with the loss function but rather with a secondary measurement, such as classification accuracy in a classification task. It is implicitly expected that the classification accuracy will be inversely proportional to the average loss value. However, in practice we often observe that the validation loss increases while the validation accuracy is stable or still improving, as illustrated by the example in Fig.~\ref{fig:Introductory_Example}. 

The cause of this behaviour can easily be thought to be shallow local optima or borderline cases of correct classification. While this explanation is consistent with classical ideas of overfitting, it does not fully explain observed behaviour. Specifically, if this is the extent of the phenomenon, there is no reason for improvement in validation accuracy if a local optimum is found, and an obvious quantitative limit to the amount that the validation loss can increase.

By investigating the distribution of per-sample validation loss values and not just a point estimation (typically averaged over all samples) we show that the increase in average validation loss can be attributed to a minority of validation samples. This means that the discrepancy between the validation loss and accuracy is due to a form of \textit{overfitting} that only affects the predictions of some validation samples, thereby allowing the model to still generalise well to most of the validation set. 

The following is a summary of the main contributions of this paper:
\begin{itemize}
    \item We present empirical evidence of a characteristic of empirical risk minimisation in MLPs performing classification tasks, that sheds light on an apparently paradoxical relationship between validation loss and classification accuracy.
    \item We explain how this phenomenon is largely a result of quantitative increases in related parameter values and the limits of using a point estimator to measure overfitting.
    \item We discuss the practical and theoretical implications of this phenomenon with regards to generalisation in related machine learning models.
\end{itemize}

In the following section we discuss related work. In Section~\ref{sec:approach} we explain our experimental setup and methodology. Section~\ref{sec:results} presents our empirical results and their interpretation. The final section  discusses and summarizes our findings with a focus on their implications for generalisation.

\section{Background}
\label{sec:background}

Much work has been done to characterize how a neural network's performance changes over training iterations \cite{Wilson2003TheGI,Neyshabur2015InSO,Hardt2016TrainFG,Hoffer2017TrainLG}. Such work has lead to some powerful machine learning techniques, including drop-out~\cite{Hinton2012ImprovingNN} and batch normalisation~\cite{Ioffe2015BatchNA}.
While both theoretically principled and practically useful generalisation bounds remain out of reach, many heuristics have been found that appear to indicate whether a trained neural network will generalise well. These heuristics have varying degrees of complexity, generality, and popularity, and include: small weight norms, flatness of loss landscapes~\cite{Hochreiter1997FlatM}, heavy-tailed weight matrices~\cite{Martin2018ImplicitSI}, and large margin distributions~\cite{Sokolic2017RobustLM}. All of these proposed metrics have empirical evidence to support their claims of contributing to the generalisation ability of a network, however, none of them have been proven to be a sufficient condition to ensure generalisation in general circumstances.

A popular experimental framework used to investigate generalisation in deep learning is to explore the optimisation process of so-called ``toy problems". Such experiments are typically characterized by varying different design choices or training conditions, in an often simplified machine learning model, and then interpreting the performance of resulting models on test data \cite{Novak2018SensitivityAG,Goodfellow2015QualitativelyCN}. The performance can be investigated post-training but it is often informative to observe how the generalisation changes \textit{during} training.

A good example of why it is important to consider performance during training is the \textit{double descent phenomenon} \cite{Belkin2018ReconcilingMM,Nakkiran2020DeepDD}. This phenomenon has enjoyed much attention recently \cite{Ba2020generalisationOT,dAscoli2020DoubleTI,Nakkiran2020OptimalRC}, due to its apparent bridging of classical and modern regimes of representational complexity. In its most basic form it is characterized by poor generalisation within a ``critically parameterized" regime of representational capacities near the minimum that is necessary to interpolate the entire training set. Slightly smaller or larger models produce improved generalisation. However, if early stopping is used the phenomenon has been found to be almost non-existent \cite{Nakkiran2020DeepDD}. 

Having an accurate estimate of test loss and how it changes during training is clearly beneficial in investigating generalisation. In the current work we show that averaging over all test samples can result in a misrepresentation of generalisation ability and that this can account for the sometimes paradoxical relationship between test accuracy and test loss.

\section{Approach}
\label{sec:approach}

We use a simple experimental setup to explore the validation loss behaviour of various fully-connected feedforward networks. All models use a multilayer perceptron (MLP) architecture where hidden layers have an equal number of ReLU-activated nodes. This architecture, while simple, still uses the fundamental principles common to many deep learning models, that is, a set of hidden layers optimised by gradient descent, using backpropagation to calculate the gradient of a given loss function with regard to the parameters being optimised.

We first determine whether the studied phenomenon (both validation accuracy and loss displaying an increase during training) occurs in general circumstances, and then select a few models where this phenomenon is clearly visible. We then probe these models to better understand the mechanism causing this effect. 

The experiments are performed on the well-known MNIST~\cite{LBBH98} and FMNIST~\cite{XRV17} classification datasets. These datasets consist of \num{60 000} training samples and \num{10 000} test samples of $28 \times 28$ grayscale images with an associated label $\in [0, 9]$. FMNIST can be regarded as a slightly more complex drop-in replacement for MNIST. Recently these datasets have become less useful as benchmarks, but they are still popular resources for investigating theoretical principles of DNNs.

All models are optimised to reduce a cross-entropy loss function measured on mini-batches of training samples. Techniques that could have a regularizing effect on the optimisation process (such as batch normalisation, drop-out, early-stopping or weight decay) were omitted as far as possible. Networks are trained till convergence, with the exact stopping criteria different for the separate experiments, as described per set of results.

A selection of hyperparameters were investigated to ensure a variety of validation loss behaviours during training. These hyperparameters are:
\begin{itemize}
    \item Training and validation set sizes;
    \item The number of hidden layers;
    \item The number of nodes in each hidden layer;
    \item Mini-batch sizes;
    \item Datasets (MNIST or FMNIST); and
    \item optimisers (Adam or SGD).
\end{itemize}
Parameter settings differed per experiment, as detailed below. Take note that the validation sets are held out from the train set, so a larger train set will result in a smaller validation set and vice versa.

\section{Results}
\label{sec:results}

Our initial experiments show that the average validation loss can indeed increase with a stable or increasing validation accuracy for a wide variety of hyperparameters (Section~\ref{subsec:increasing_risk}). Based on this result, we select a few models where the phenomenon is clearly visible, and investigate the per-sample loss distributions throughout training, as well as weight distributions, to probe the reason for this behaviour (Sections \ref{subsec:loss_distributions} and \ref{subsec:val_set_outliers}).

\subsection{Increasing Risk During Training}
\label{subsec:increasing_risk}

We begin our investigation by training $95$ two-layer MLPs, varying the width of the hidden layer and the optimisation algorithm over multiple random initialisations. Networks are trained on $5\ 000$ MNIST training samples using a mini-batch size of $64$. All models were trained until interpolation (training accuracy of $100\%$), which occurred at around $3\ 000$ epochs for the smaller models. Out of the $95$ models trained, $57$ ($3$ initialisations of $19$ widths) were optimised with Adam and the remaining $38$ ($2$ initialisations of $19$ widths) with SGD.

\begin{figure}[!h]
        \centering
        \includegraphics[width=0.95\linewidth]{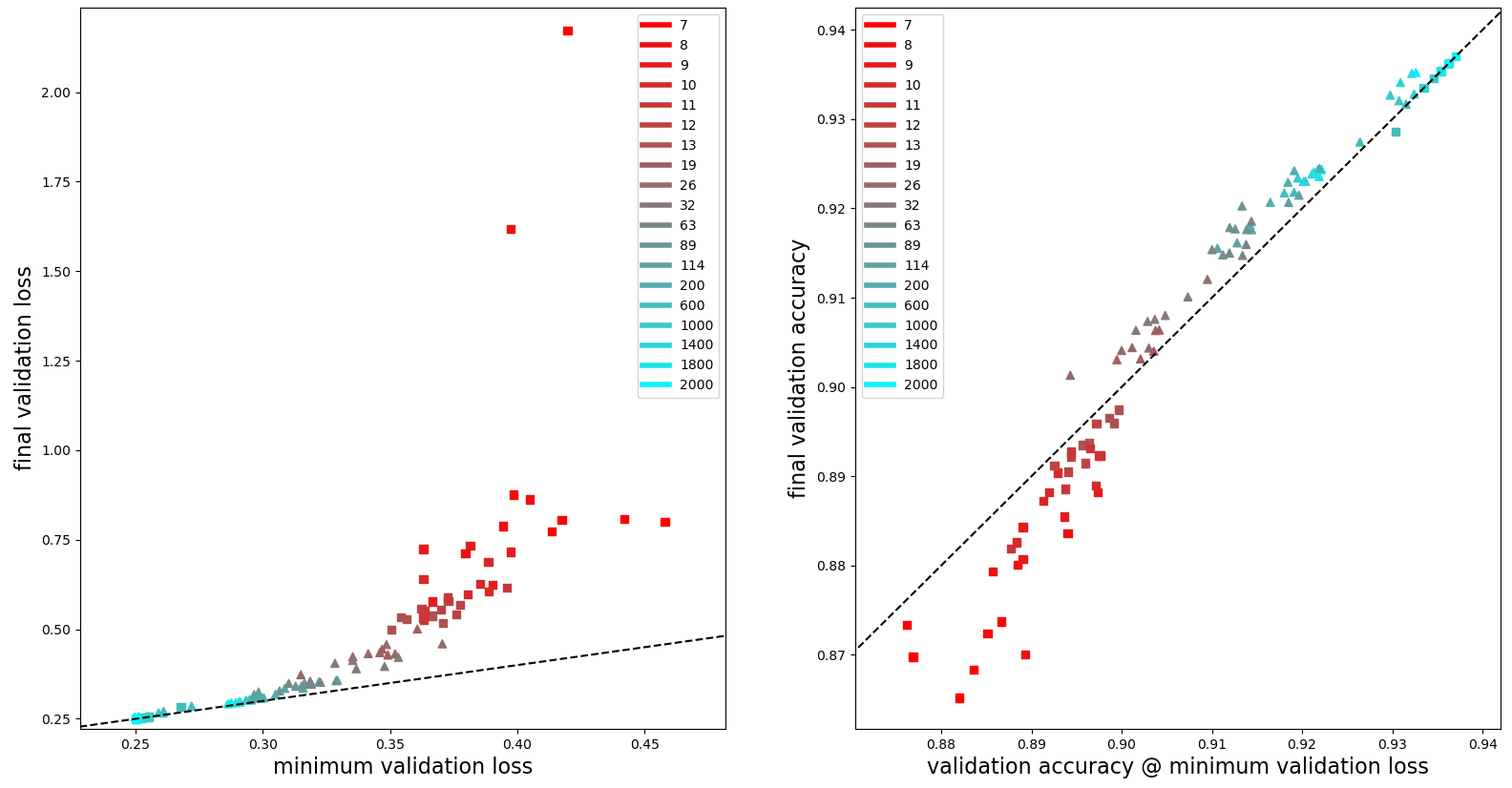}
        \caption{Final validation loss (left) and validation accuracy (right) vs the same metric at the epoch with {\em minimum} validation loss. $95$ MLPs are trained on $5\ 000$ MNIST samples with the number of nodes in the hidden layer ranging from only $7$ (red) to $2\ 000$ (blue). Models marked with a triangle had increasing validation loss and accuracy between the epoch of minimum validation loss and the epoch of first interpolation.}
        \label{fig:1xN scatter}
\end{figure}

The results are presented in the scatter plots in Fig.~\ref{fig:1xN scatter}. The measurement for the horizontal axis is made at the epoch where the model achieved the lowest validation loss. The measurement for the vertical axis is made at the epoch where the model first interpolated the entire training set. Using the linear curve as reference, all models falling above the line increased the relevant metric after the point of minimum validation loss. The models marked by a triangle saw increases in both validation loss and accuracy.

Notice that the models with limited representational capacity, in this case referring to the number of nodes in the hidden layer, had increasing validation loss and decreasing validation accuracy as one would expect. This is in contrast with the larger models that tend to display an increase in both validation accuracy and loss even before interpolation. An additional observation we can make is that a higher minimum validation loss seems to be indicative of very large increases in validation loss later in training. 

\begin{figure}[!h]
        \centering
        \includegraphics[width=0.95\linewidth]{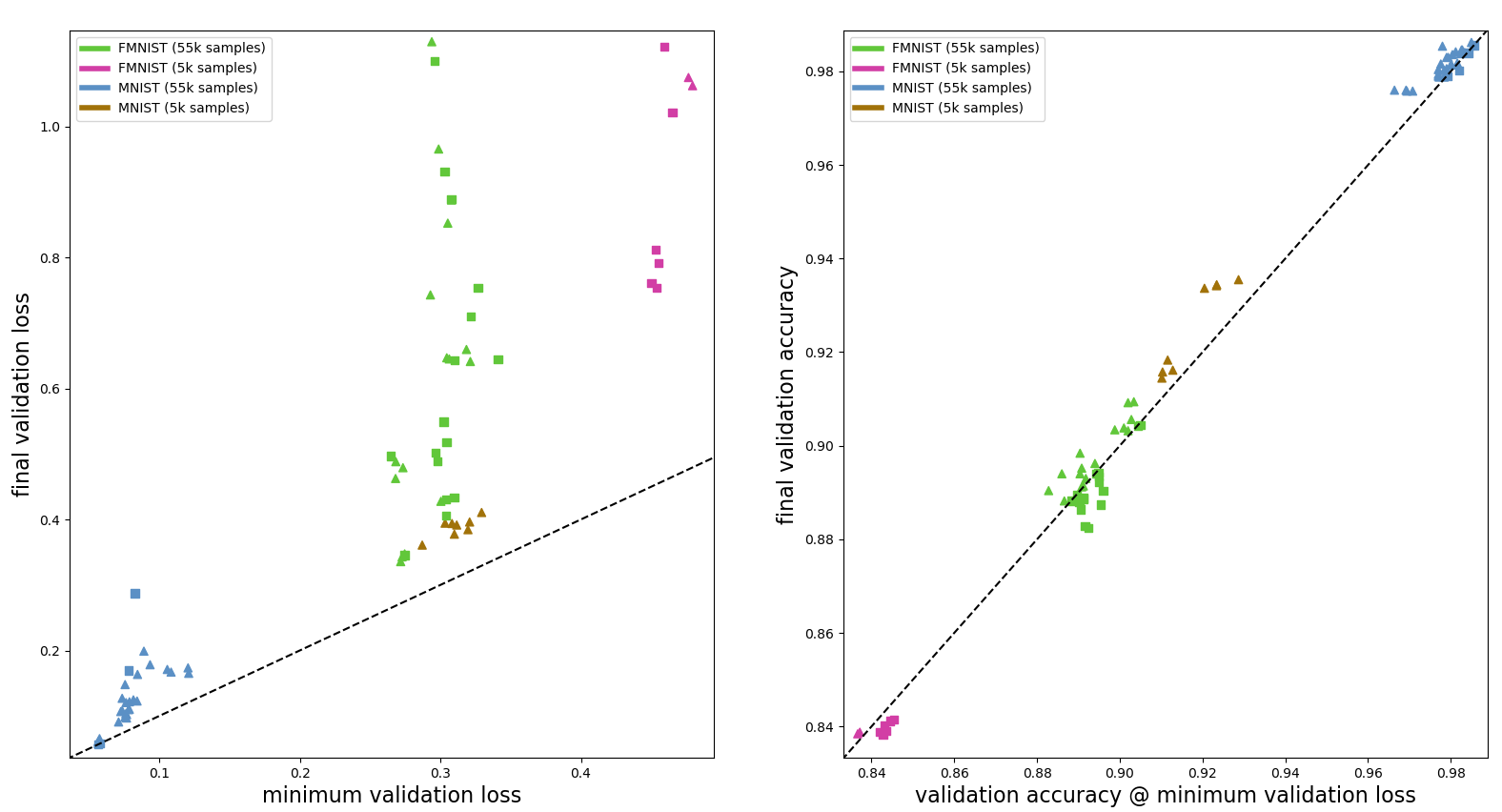}
        \caption{Final validation loss (left) and validation accuracy (right) vs the same metric at the epoch with {\em minimum} validation loss. $80$ MLPs are trained with varying hyperparameters; colours refer to different training sets. Models marked with a triangle had increasing validation loss and accuracy between the epoch of minimum validation loss and the epoch of first interpolation.}
        \label{fig:main scatter}
\end{figure}

We repeated this experiment in a more general setting to produce the results presented in Fig.~\ref{fig:main scatter}. These models have hyperparameters that vary beyond just the hidden layer width and optimiser. Specifically:
\begin{itemize}
    \item The data is either MNIST or FMNIST with a train set size as defined in the legend;
    \item The number of hidden layers is either $1$, $3$, or $10$;
    \item The number of nodes in each hidden layer is either $100$ or $1 000$;
    \item Mini-batch sizes are either $16$, $64$ or $256$; and
    \item The optimisers are again either Adam or SGD.
\end{itemize}

These networks are trained for 150 epochs. In order to ensure that each model's performance is good enough to be considered typical for these architectures and datasets, we use optimised learning rates. The learning rate for each set of hyperparameters is chosen by a grid search over a wide range of values. The selection is made in accordance with the best validation error achieved by the end of training. In some cases this resulted in final training accuracies slightly below $100\%$. In these cases we selected the ``final" epoch at the epoch where maximum training accuracy was achieved.

As expected, the models trained on MNIST or on larger training sets had lower validation losses and higher validation accuracies in general. However, we also note that models trained on FMNIST tend to have much higher increases in validation loss while the validation accuracy is still improving when compared to models trained on MNIST. In the next section we investigate how the loss distributions change for selected models from this section.

\subsection{Loss Distributions}
\label{subsec:loss_distributions}


In order to take a closer look at how both loss and accuracy can increase during training, we present a case study of four selected models from the previous section. They are defined below.

\begin{itemize}
    \item A:    3x100 model trained on $5\ 000$ MNIST samples using Adam and a mini-batch size of 64.
    \item B:    3x100 model trained on $5\ 000$ MNIST samples using SGD and a mini-batch size of 64.
    \item C:    3x100 model trained on $5\ 000$ FMNIST samples using SGD and a mini-batch size of 64.
    \item D:    3x100 model trained on $55\ 000$ FMNIST samples using Adam and a mini-batch size of 16.
\end{itemize}

\begin{figure}[!h]
    \centering
    \includegraphics[width=0.98\linewidth]{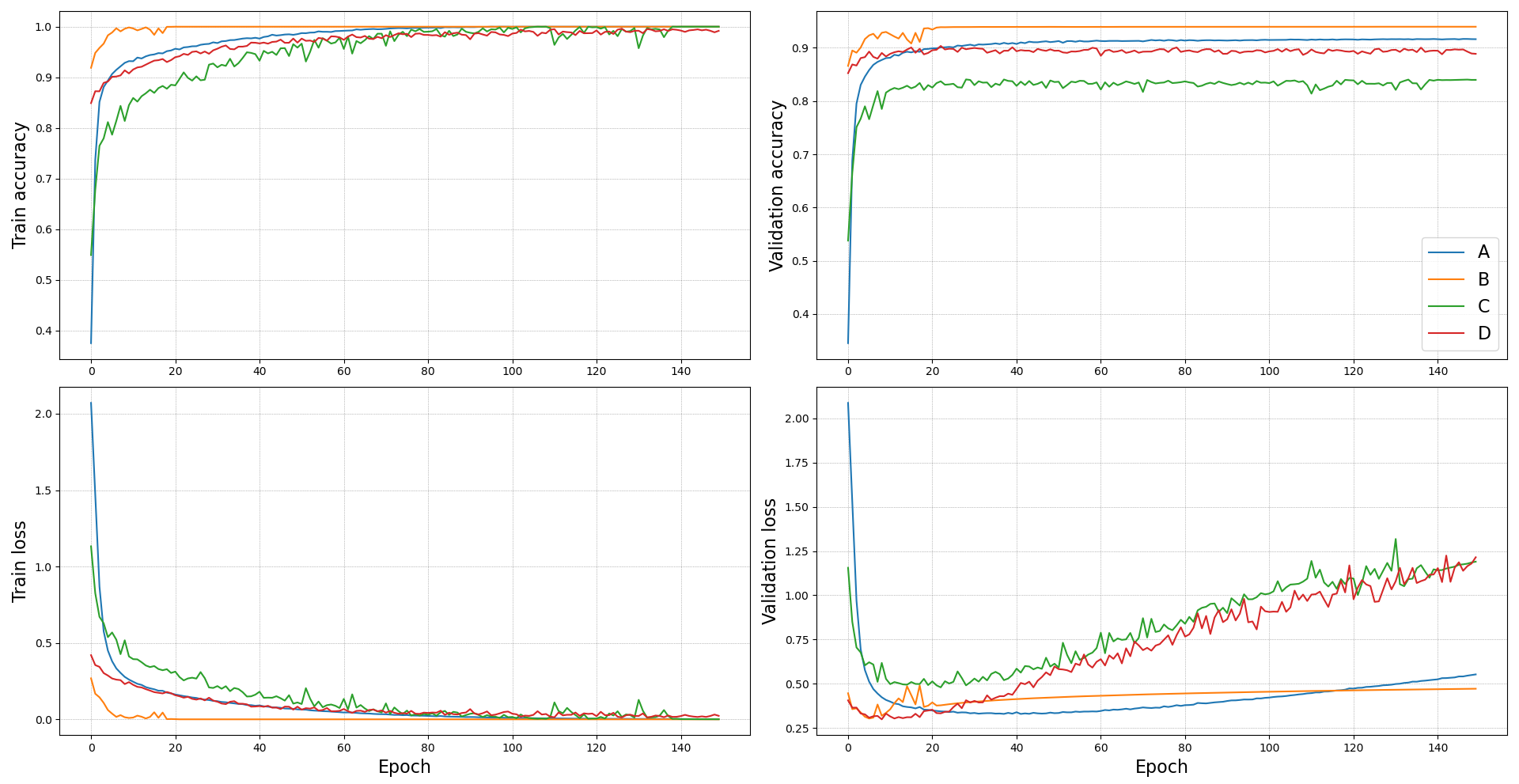}
    \caption{Learning curves for four selected models (A-D, see text) showing increasing validation loss, despite an increasing or stable validation accuracy.}
    \label{fig:selected learning curves}
\end{figure}

The learning curves for these models are presented in Fig.~\ref{fig:selected learning curves}. Notice that for all four models a minimum validation loss is achieved early on. Beyond this point the validation loss increases while the corresponding accuracy is either stable or improving slightly. 

The validation loss curve, as seen in Fig.~\ref{fig:selected learning curves}, is often used as an estimate of the level of overfitting that is occurring as the model is optimised on the training set. However, by averaging over the entire validation set we are producing a point estimate that implicitly assumes that the loss of all validation samples are close to the mean value. This assumption is reasonable with regards to the training set because most loss functions (e.g. cross entropy) work with the principle of maximum likelihood estimation \cite{murphy,GBC16}. This means that by minimizing the dissimilarity between the entire distribution of the training data and the model the estimate of loss is all but guaranteed to be indicative of performance on the entire training set.

There is no such guarantee with regards to any set other than the training set. This makes the average loss value a poor estimate of performance on the validation set. The results presented in Fig.~\ref{fig:SO_Loss} motivate this point for model $A$. See Appendix.~\ref{appendix: Loss and weight distributions} for the same results for models $B$, $C$, and $D$. The plots show heatmaps of loss distributions for the four selected models at several training iterations for three datasets (training, validation and evaluation). The validation set is the held-out set that is used to estimate performance during training and model selection, and the evaluation set is the set that is used post-training to ensure no indirect optimisation is performed on the test data. The iterations refer to parameter updates, not epochs. We show the distributions at log-sampled iterations because many changes occur early on (even before the end of the first epoch) and few occur towards the end of training. A final note with regards to these heatmaps is that the colours, which define the number of samples that have the corresponding loss value, are also log-scaled. This visually highlights the occurrence of samples with extreme loss values.

The loss distributions in Fig.~\ref{fig:SO_Loss} show that while the loss value for a vast majority (indicated by the red and orange colours) of samples reduces with training iterations there is a small minority of samples for which the loss values increase. For the training set, this increase is relatively low and eventually reduces as the entire set is interpolated. For the validation and evaluation sets the loss values of these ``outliers" seem to only increase. This is why it is possible for the average validation loss to increase while the classification accuracy remains stable or improves.

\begin{figure}[!h]
    \includegraphics[width=\linewidth]{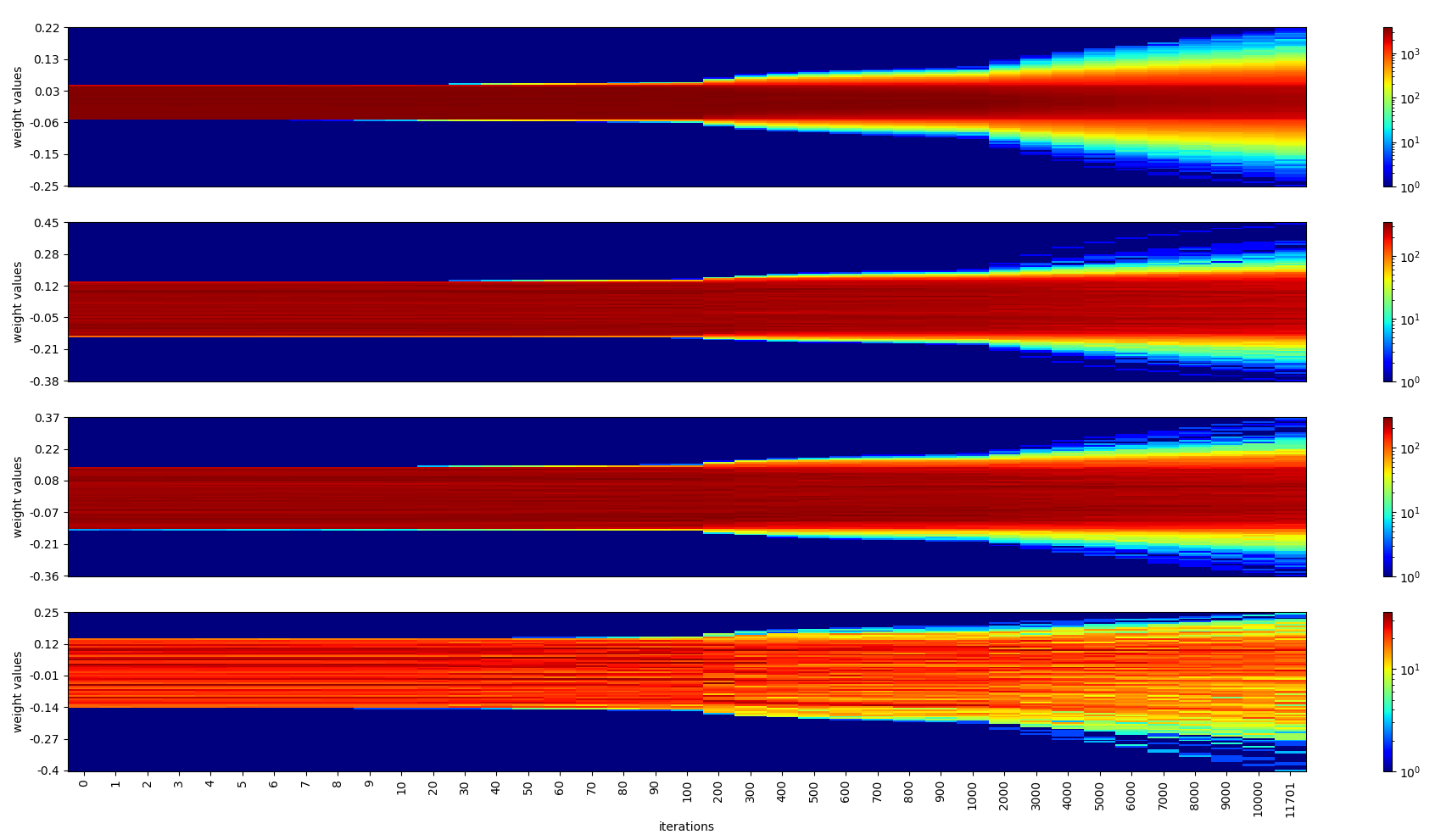}
    \caption{Change in weight distributions during training for model A (MNIST, Adam, mini-batch size of 64). Each heatmap refers to a layer in the network, including the output layer, from top to bottom.}
    \label{fig:SO_Weights}
\end{figure}

Fig.~\ref{fig:SO_Weights} shows the weight distributions for the same model in the same format as Fig.~\ref{fig:SO_Loss}. See Appendix.~\ref{appendix: Loss and weight distributions} for models $B$, $C$, and $D$. It can be observed that there is a clear increase in the magnitude of some weights (their absolute weight values) at the same iterations where we observe a corresponding increase in validation and evaluation loss values in Fig.~\ref{fig:SO_Loss}. This appears to occur even more after most of the training sample losses have been minimized. This is consistent with the notion of limiting weight norms to improve generalisation and it suggests that the reason for the increase in validation losses is because particular weights are being increased to fit idiosyncratic training samples.

\begin{figure}[!h]
    \includegraphics[width=\linewidth]{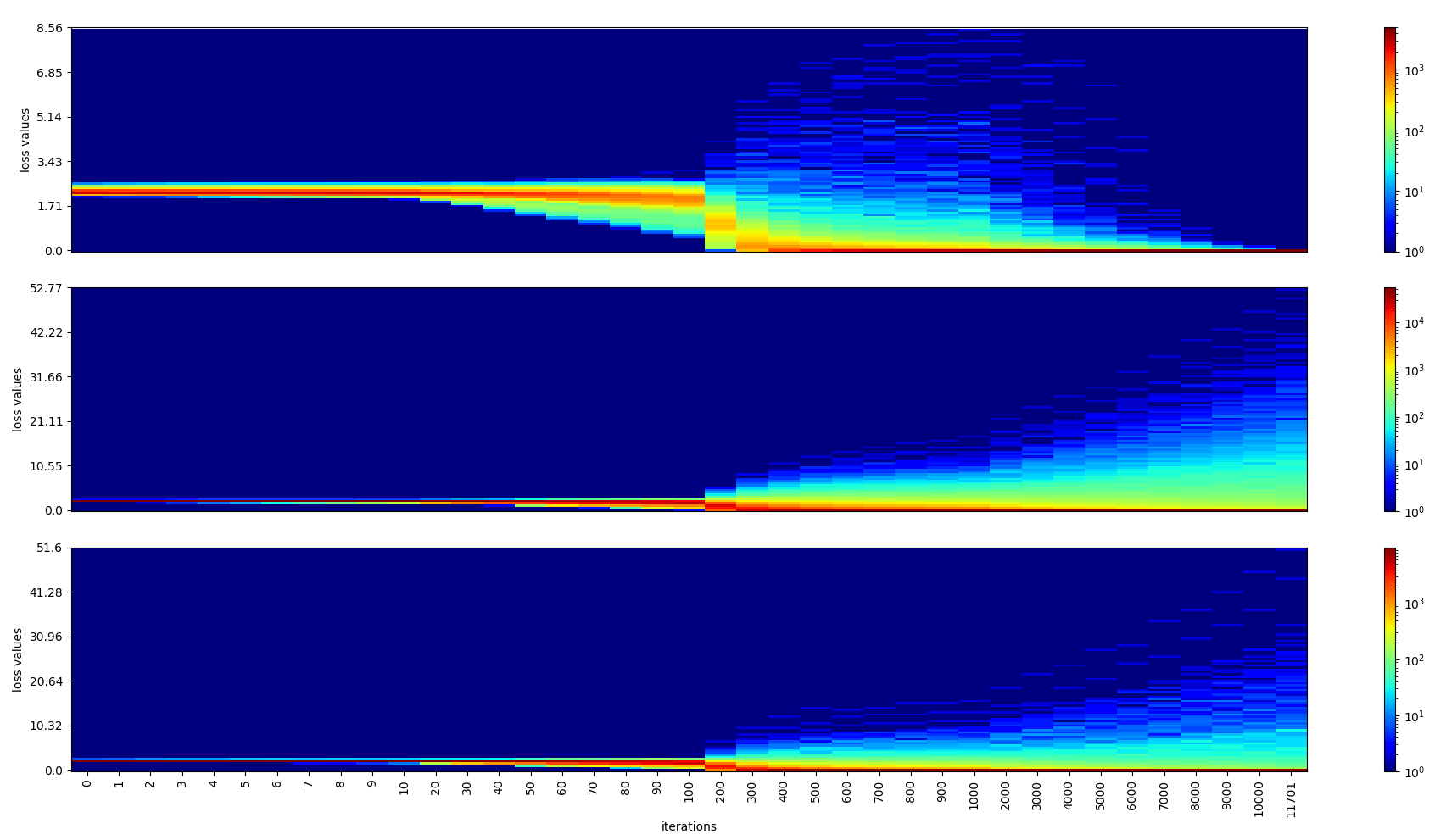}
    \caption{Change in loss distributions during training for model A (5k MNIST, Adam, mini-batch size of 64). The three heatmaps refer to the train (top), validation (center), and evaluation (bottom) loss distributions.}
        \label{fig:SO_Loss}
\end{figure}

While these heatmaps show that there are \textit{outlier} per-sample loss values in the validation set, they do not guarantee that these extreme loss values are due to specific samples. It is possible that the extreme values are measured on completely different samples at every measured iteration, in which case there is nothing extreme about them and the phenomenon has something to do with the optimisation process and not training and validation distributions. We address this question in the next section.

\subsection{Validation Set Outliers}
\label{subsec:val_set_outliers}

In this section we investigate whether the validation set samples with extreme loss values are individual samples that are consistently modeled poorly, or whether these outliers change from iteration to iteration due to the stochastic nature of the optimisation process. Towards this end, we analyze the number of epochs for which a sample can be regarded as an outlier and compare it with its final loss value. 

We classify a sample as an outlier when its loss value is above the upper Tukey fence, that is, larger than $Q_{3} + 1.5 \times (Q_{3} - Q_{1})$, where $Q_{1}$ and $Q_{3}$ are the first and third quartile of all loss values in the validation set, respectively~\cite{devore2005applied}. This indicator is simple and adequate to illustrate whether some specific samples consistently have much larger loss values than the majority.

\begin{figure}[!h]
    \centerline{
        \begin{subfigure}[t]{.5\textwidth}
            \centering
            \includegraphics[width=0.98\linewidth]{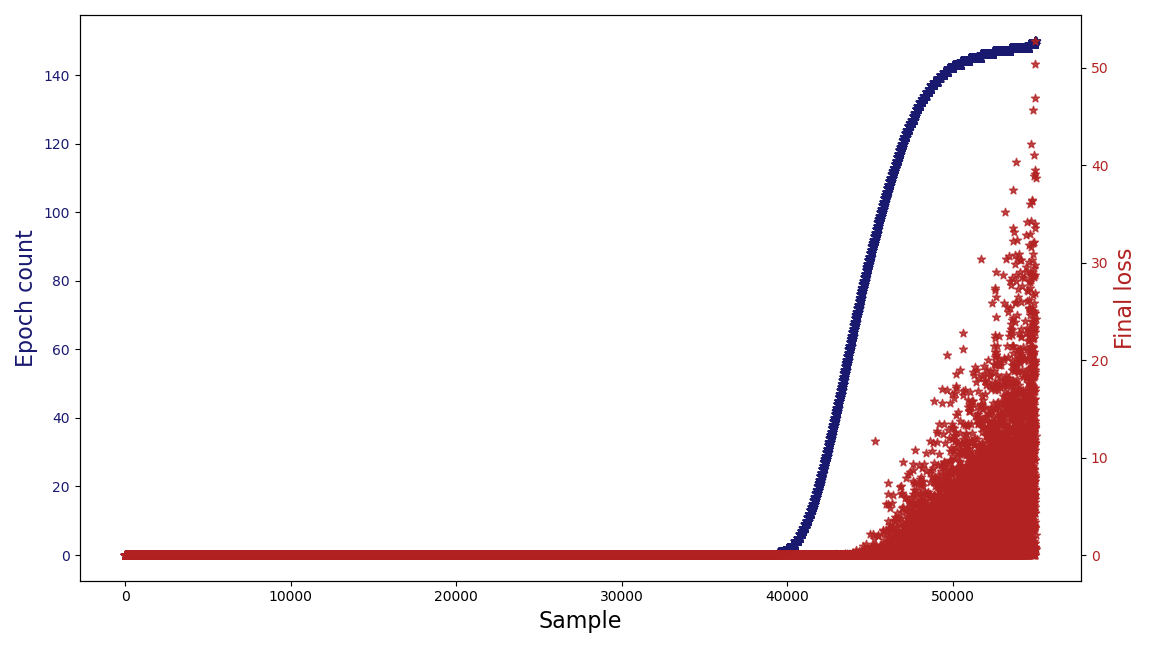}
            \caption{A: Fitting 5k MNIST; Adam}
            \label{subfig:SO_Adam_MNIST_outliers}
        \end{subfigure}
    
        \begin{subfigure}[t]{.5\textwidth}
            \centering
            \includegraphics[width=0.98\linewidth]{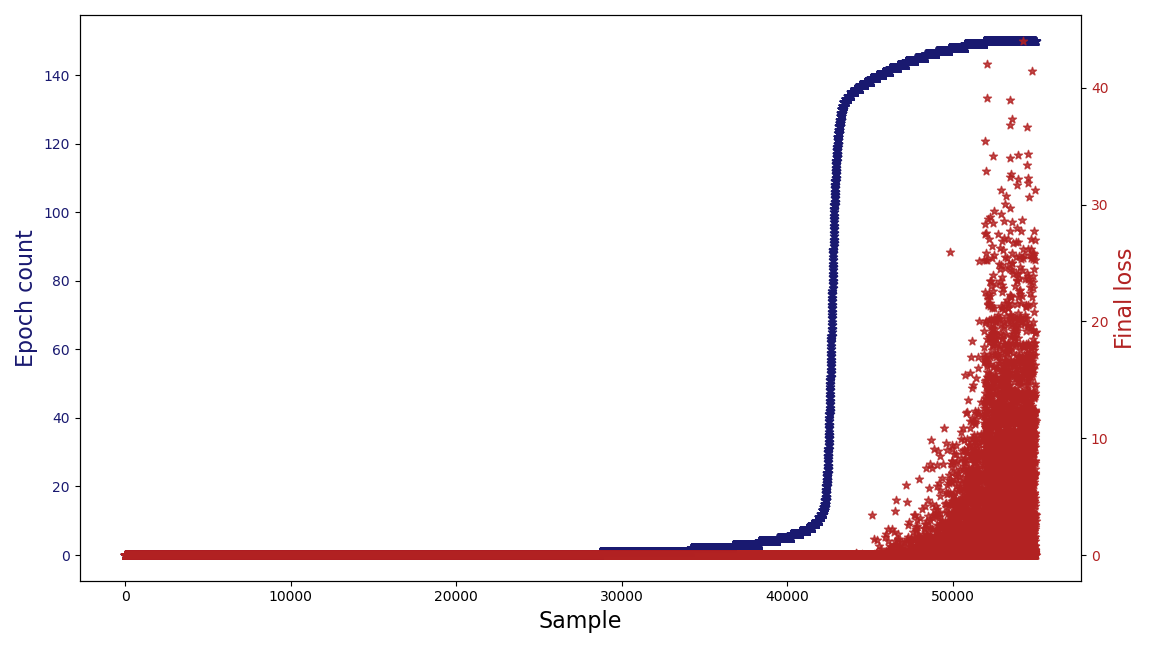}
            \caption{B:  Fitting 5k MNIST; SGD}
            \label{subfig:SO_SGD_MNIST_outliers}
        \end{subfigure}
    }
    \centerline{
        \begin{subfigure}[t]{.5\textwidth}
            \centering
            \includegraphics[width=0.98\linewidth]{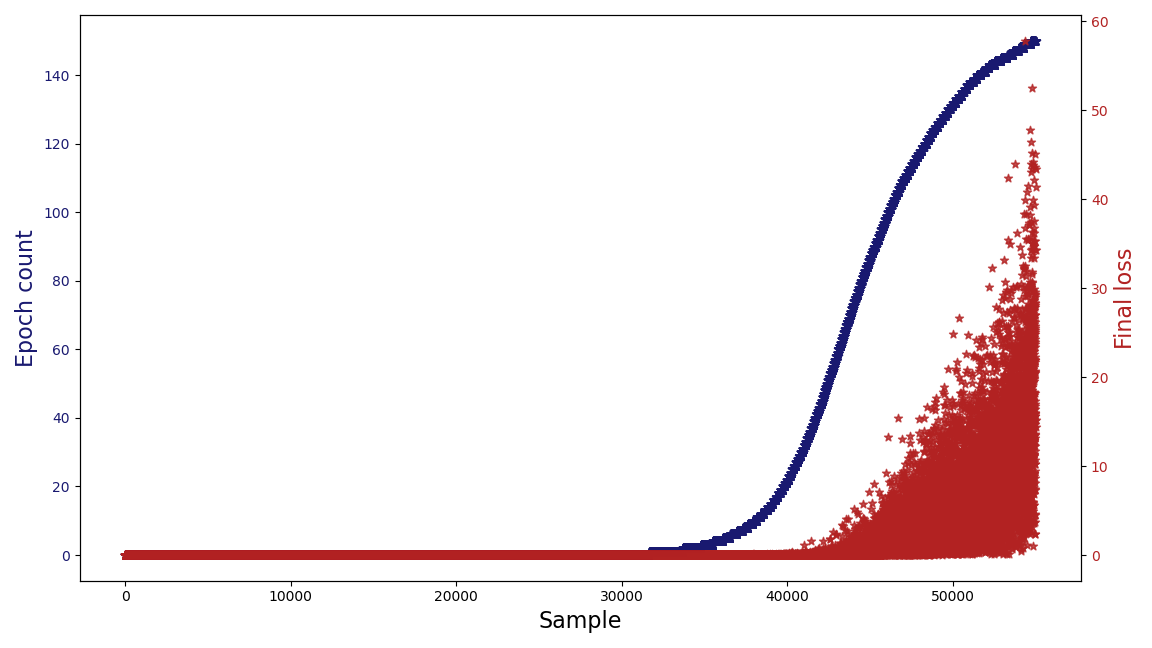}
            \caption{C:  Fitting 5k FMNIST; SGD}
            \label{subfig:SO_SGD_FMNIST_outliers}
        \end{subfigure}
        
        \begin{subfigure}[t]{.5\textwidth}
            \centering
            \includegraphics[width=0.98\linewidth]{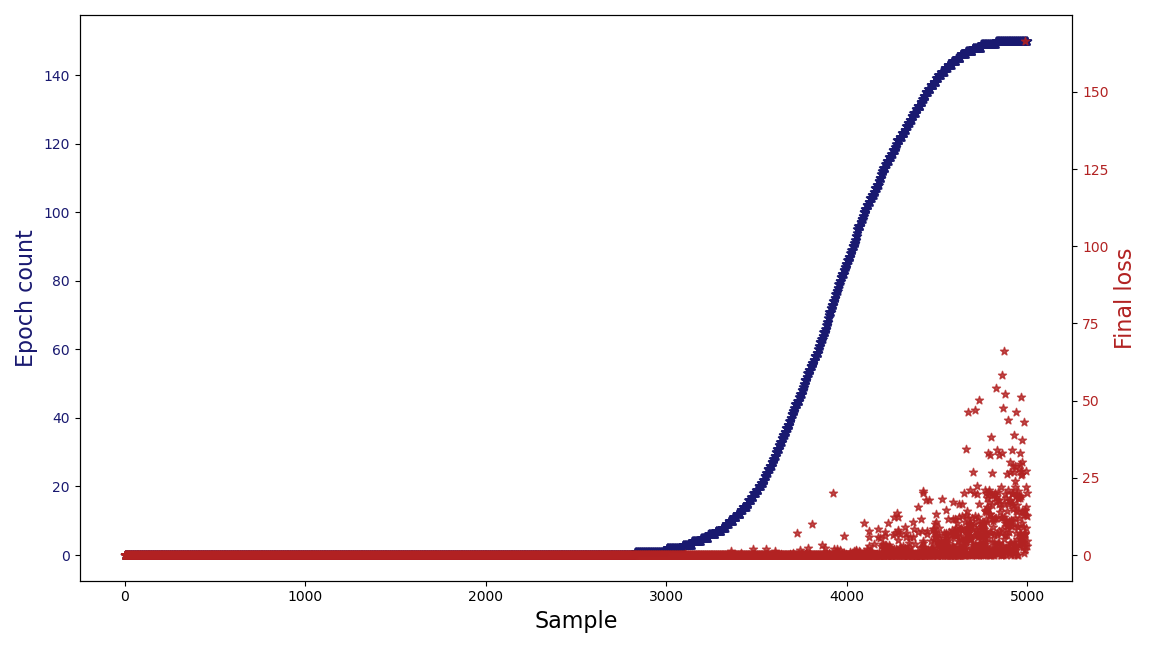}
            \caption{D:  Fitting 55k FMNIST; Adam}
            \label{subfig:SO_Adam_FMNIST_outliers}
        \end{subfigure}\\
    }
    
    \caption{Outliers in the validation set. The blue datapoints show the number of epochs for which each sample is considered an outlier. The red datapoints show the loss value of each sample at the end of training. Samples are ordered in ascending order of epoch counts.}
    \label{fig:ISE}
\end{figure}

In Fig.~\ref{fig:ISE} we show that the validation samples with extreme loss values at the end of training are usually classified as outliers for most of the training process. This means that the extreme validation loss values are due to specific samples that are not well modeled. In addition to this, it is worth observing that a large majority of validation samples are never classified as an outlier and these samples always have small loss values at the end of training.

\section{Discussion}
\label{sec:discussion}

We have shown that validation classification accuracy can increase while the corresponding average loss value also increases. Empirically, we have noted that this phenomenon is most influenced by the interplay between the training dataset and model capacity. Specifically, it occurs more for larger models, smaller training datasets, and more difficult datasets (FMNIST in our investigation). We can, however, combine the first and second factors because capacity is directly related to the size and complexity of the training set.

By taking a closer look at per-sample loss distributions and weight distributions we have noted that the phenomenon is largely due to specific samples in the validation set that have extremely large loss values and obtain progressively larger loss values as training continues. These loss values then become large enough to distort the average loss value in such a way that it appears that the model is overfitting the training set, when most of the validation set sample losses are still being minimized. From a theoretical viewpoint this is unsurprising because the average validation loss is only a good measure of risk with regards to the train set, where it is directly being minimized by the principle of maximum likelihood estimation. From a practical viewpoint it appears that increased weight values are sacrificing the generality of the distributed representation used by DNNs in order to minimize training loss as much as possible.

Practically, these findings serve as a clear cautionary tale for (1) assuming an inverse correlation between loss and accuracy, and for (2) measuring overfitting with point estimators such as average validation loss. Rather, we show that loss distribution heatmaps (Fig. \ref{fig:SO_Loss}) provide additional, useful information.  

The findings also highlight a more general aspect of generalisation and deep learning: DNNs optimise parameters with regards to training data in a heterogeneous manner. With sufficient parametric flexibility, these types of models can fit generalisable features and memorize non-generalisable features concurrently during training. Formally defining how this is achieved, and subsequently, how generalisation should be characterized in this context, remains an open problem.

\section{Conclusion}
\label{sec:conclusion}

By means of a small but focused empirical investigation we have contributed the following findings, in the context of using fully-connected feedforward networks as classifiers:

\begin{itemize}
    \item If the representational capacity is large enough, validation classification accuracy and loss can both increase simultaneously during training.
    \item Under common conditions, average validation loss is a poor estimate of generality because validation samples are not guaranteed to obtain loss values near the mean value.
    \item We show that sample-specific heatmaps provide a far more nuanced view of the training process, and can be a useful tool during model optimisation.
    \item We propose that investigations of generalisation should consider the fact that DNN optimisation is distributed and heterogeneous, which is why simple measures of overfitting can be misleading.
\end{itemize}

These findings imply that a validation loss that starts increasing prior to interpolation of the training set is not necessarily an implication of overfitting; and also that it is dangerous to assume a negative correlation between validation accuracy and loss (which is often done when selecting hyperparameters). 


While this study aimed to answer a very specific question, we hope it will contribute to the general discourse on factors that influence the optimisation process and generalisation ability of neural networks. 

\newpage
\appendix

\section{Appendix}
\label{appendix: Loss and weight distributions}

We include the results when models B, C and D are analyzed, using the same process as described in Section \ref{subsec:loss_distributions}. 

\begin{figure}[!h]
    \centerline{
        \begin{subfigure}[t]{.5\textwidth}
            \includegraphics[width=\linewidth]{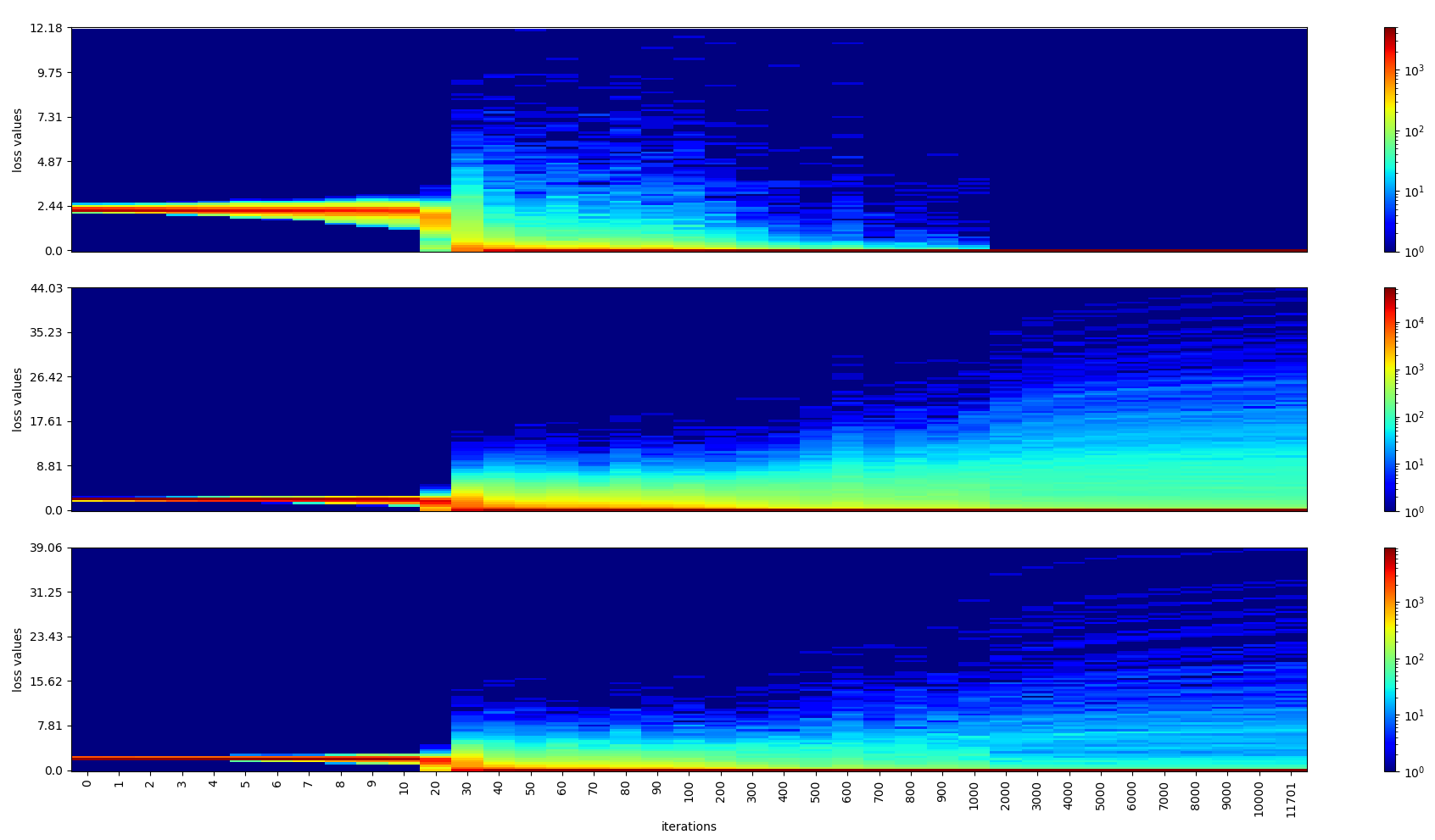}
            \caption{B:  Fitting 5k MNIST; SGD}
            \label{subfig:SO_SGD_MNIST_Loss}
        \end{subfigure}
        \begin{subfigure}[t]{.5\textwidth}
            \includegraphics[width=\linewidth]{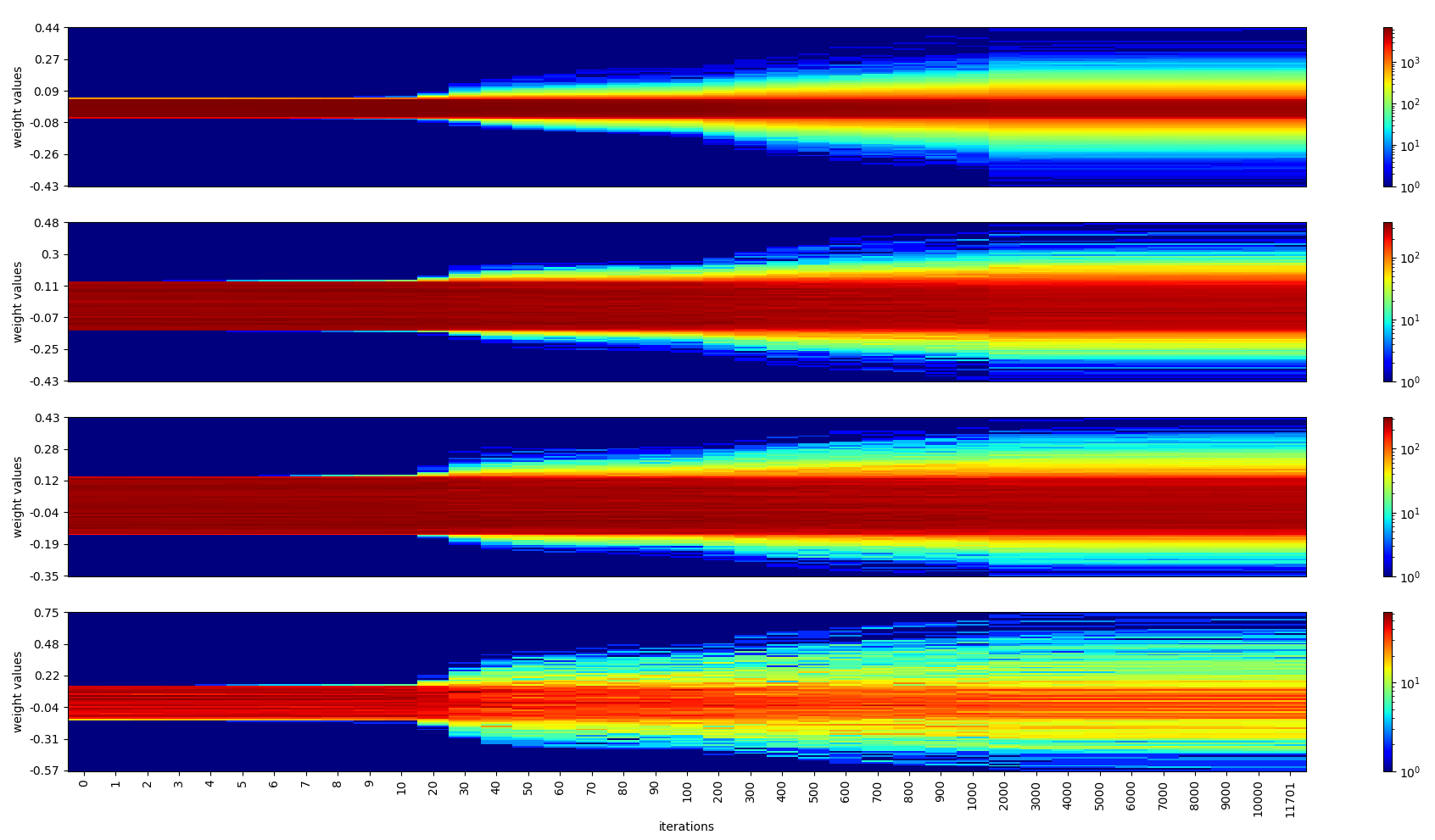}
            \caption{B:  Fitting 5k MNIST; SGD}
            \label{subfig:SO_SGD_MNIST_Weight}
        \end{subfigure}
    }
    \centerline{
        \begin{subfigure}[t]{.5\textwidth}
            \includegraphics[width=\linewidth]{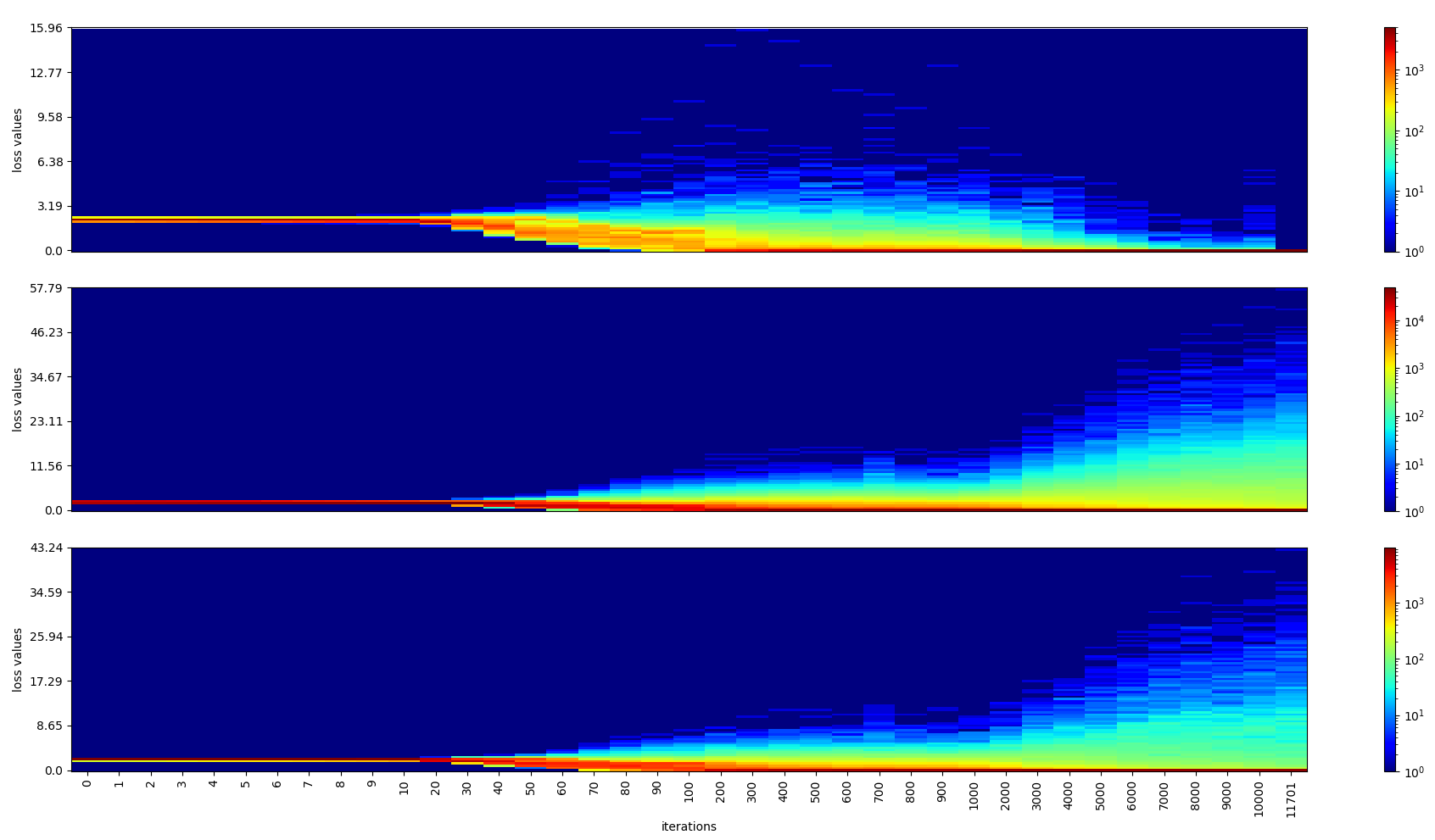}
            \caption{C:  Fitting 5k FMNIST; SGD}
            \label{subfig:SO_SGD_FMNIST_Loss}
        \end{subfigure}
        \begin{subfigure}[t]{.5\textwidth}
            \includegraphics[width=\linewidth]{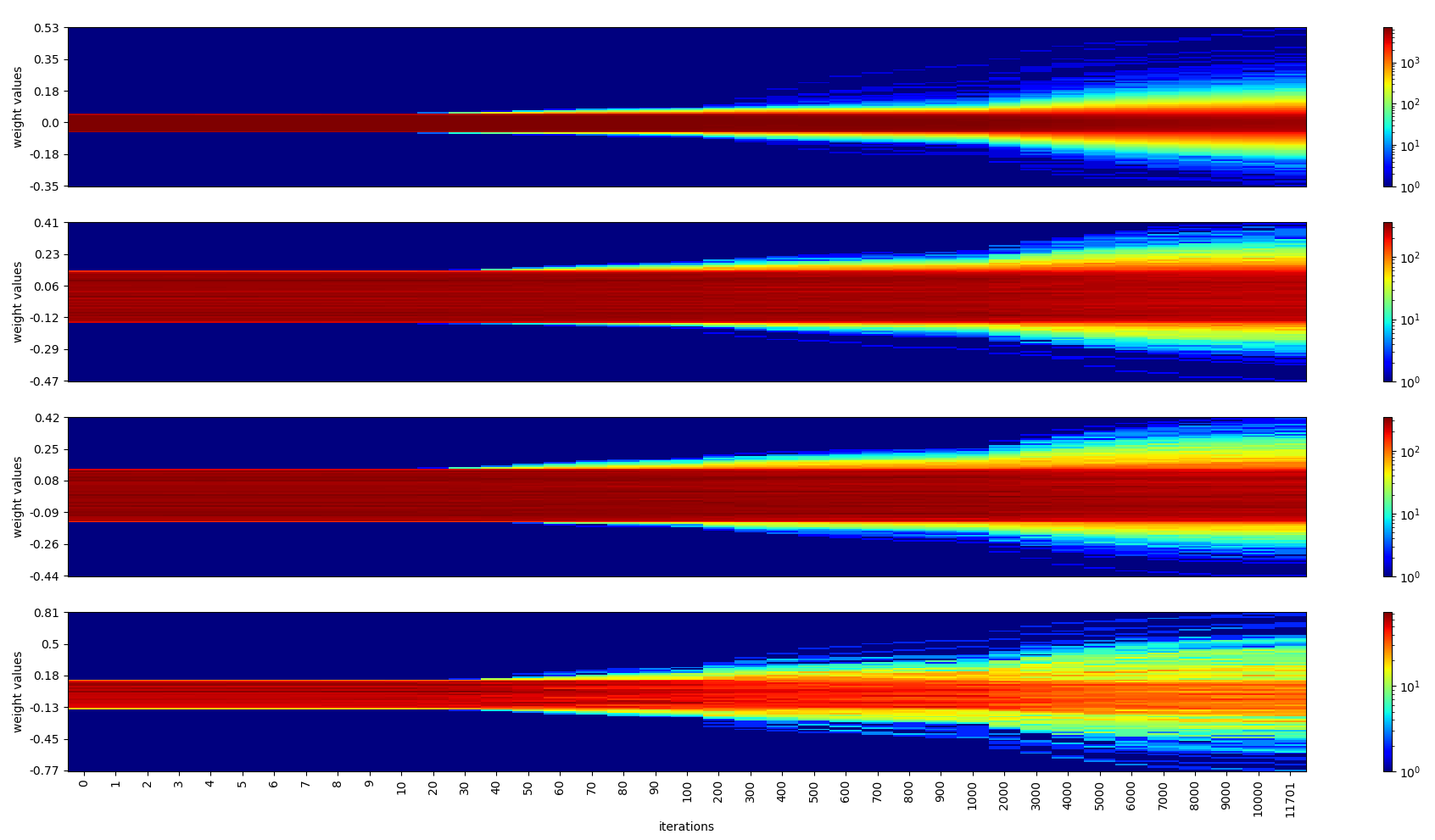}
            \caption{C:  Fitting 5k FMNIST; SGD}
            \label{subfig:SO_SGD_FMNIST_Weight}
        \end{subfigure}
    }
    \centerline{
        \begin{subfigure}[t]{.5\textwidth}
            \includegraphics[width=\linewidth]{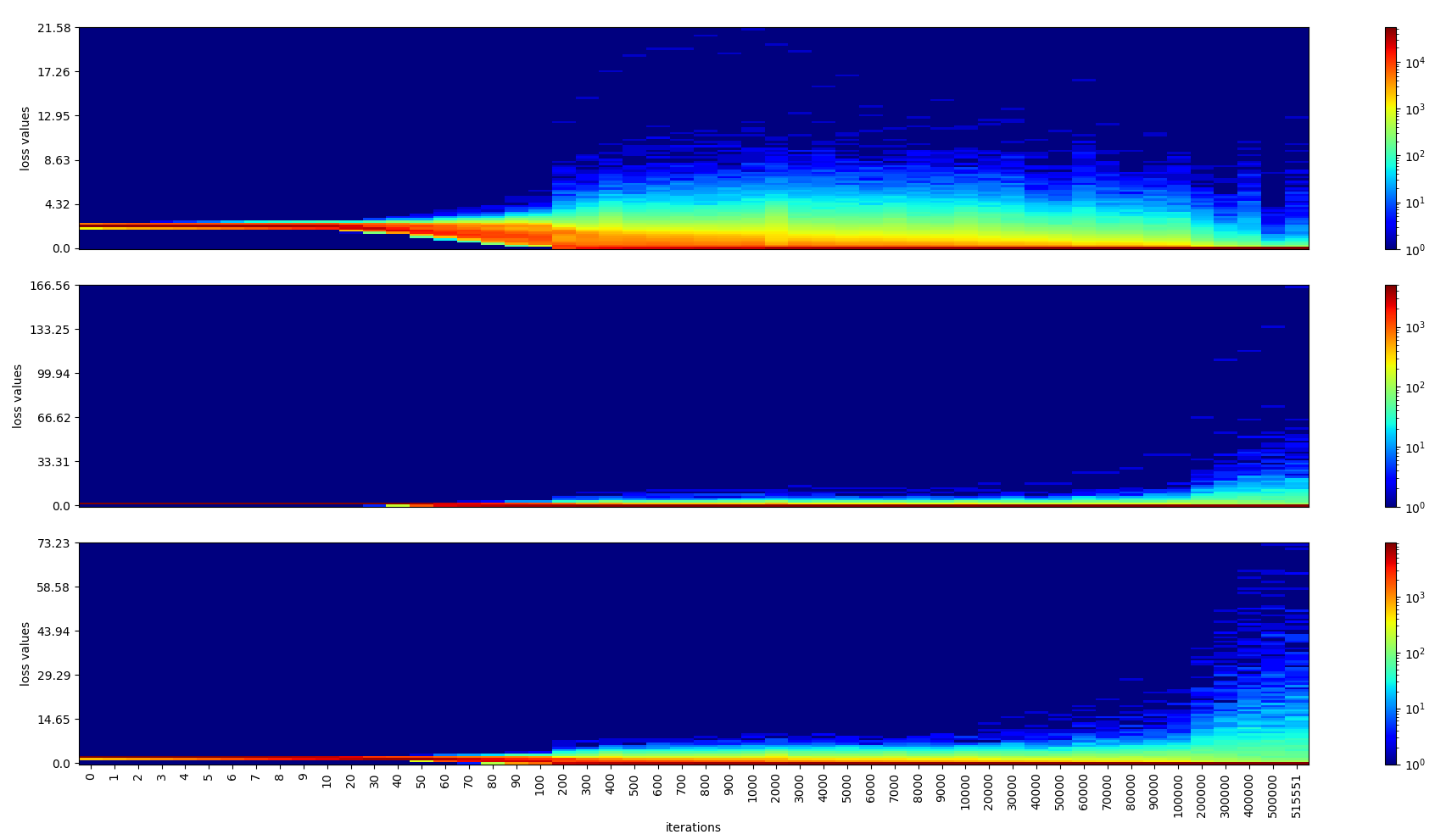}
            \caption{D:  Fitting 55k FMNIST; Adam}
            \label{subfig:SO_Adam_FMNIST_Loss}
        \end{subfigure}
        \begin{subfigure}[t]{.5\textwidth}
            \includegraphics[width=\linewidth]{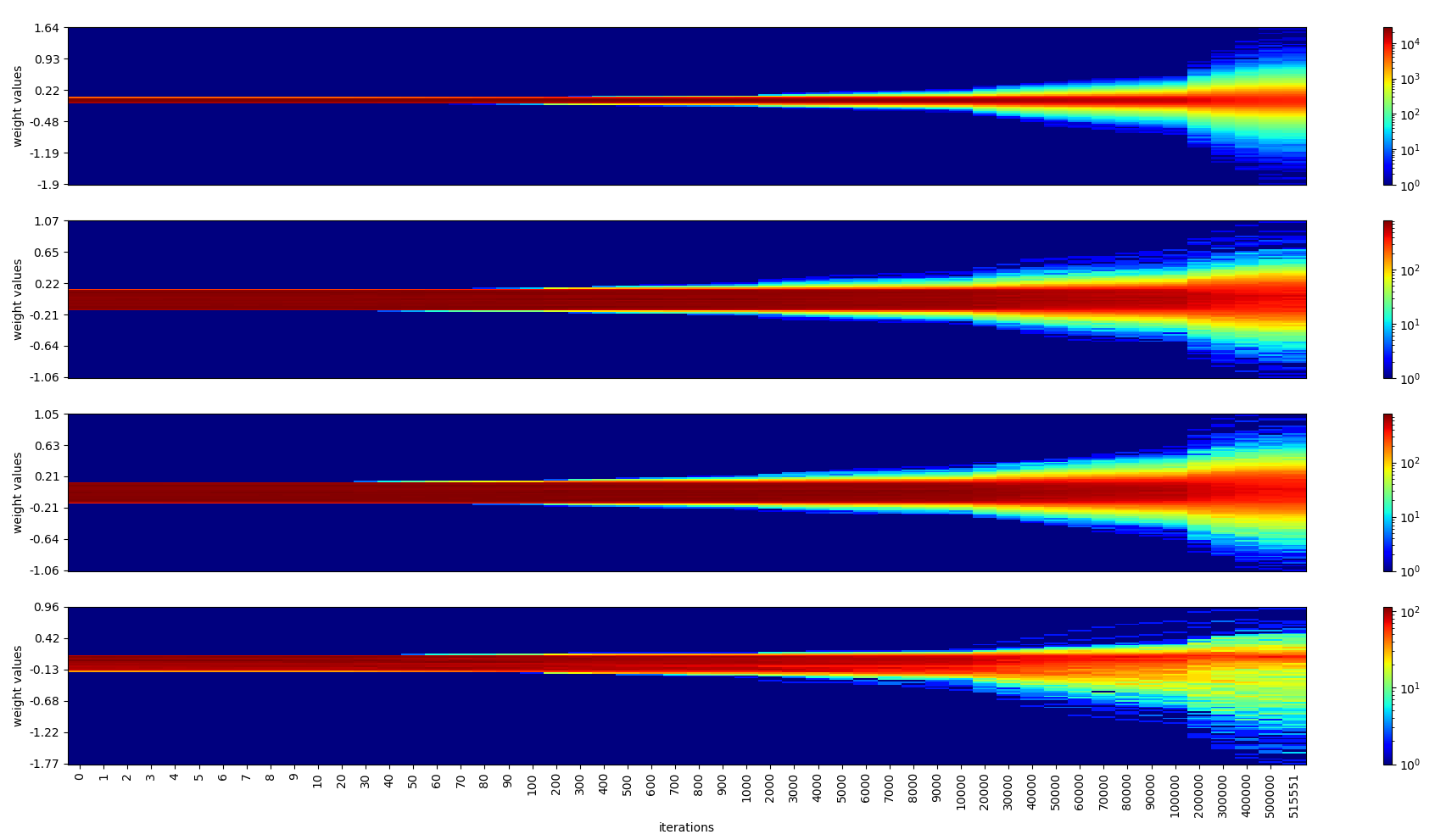}
            \caption{D:  Fitting 55k FMNIST; Adam}
            \label{subfig:SO_Adam_FMNIST_Weight}
        \end{subfigure}
    }
    \caption{Change in loss (left) and weight (right) distributions, for models B, C, and D, during training. See Fig.~\ref{fig:SO_Loss} and \ref{fig:SO_Weights} for plot ordering.}
    \label{fig:SO_Loss_appendix}
\end{figure}

\bibliographystyle{unsrt}  
\bibliography{references}  

\end{document}